\documentclass[10pt,twocolumn,letterpaper]{article}

 \usepackage{cvpr}              

\newcommand{\red}[1]{\textcolor{black}{#1}}
\newcommand{\redv}[1]{\textcolor{black}{#1}}

\usepackage{tabularx}
\usepackage{booktabs}
\usepackage{multirow}
\usepackage{pifont}
\usepackage{subcaption}
\usepackage{indentfirst}
\usepackage{threeparttable}
\usepackage{footmisc}  
\usepackage[pagebackref,breaklinks,colorlinks,allcolors=cvprblue]{hyperref}
\usepackage[accsupp]{axessibility}  

\definecolor{cvprblue}{rgb}{0.21,0.49,0.74}

\def\paperID{*****} 
\def\confName{CVPR}
\def\confYear{2025}

\title{Forecasting When to Forecast: Accelerating Diffusion Models with Confidence-Gated Taylor}

\author{Xiaoliu Guan$^{1,2}$ \and
Lielin Jiang$^{2}$ \and
Hanqi Chen$^{2,3}$ \and
Xu Zhang$^{2}$ \and
Jiaxing Yan$^{2}$ \and
Guanzhong Wang$^{2}$ \and
Yi Liu$^{2}$ \and
Zetao Zhang$^{4}$ \and
Yu Wu$^{1,*}$ \and
$^{1}$School of Computer Science, Wuhan University \\
$^{2}$PaddlePaddle Team, Baidu Inc \\
$^{3}$International Joint Innovation Center, The Electromagnetics Academy at Zhejiang University \\
$^{4}$Yunnan Key Laboratory of Media Convergence\and
{\tt\small \{liuxiaoguan, wuyucs\}@whu.edu.cn  \tt\small  22360175@zju.edu.cn \tt\small zzt@yndaily.com} \\
\tt\small \{jianglielin, zhangxu44, yanjiaxing, wangguanzhong, liuyi22\}@baidu.com 
}

\begin{document}

\renewcommand{\thefootnote}{\fnsymbol{footnote}}
\twocolumn[{
\renewcommand\twocolumn[1][]{#1}
\maketitle
\vspace{-3.5em}  
\begin{center}
    \includegraphics[width=0.32\textwidth]{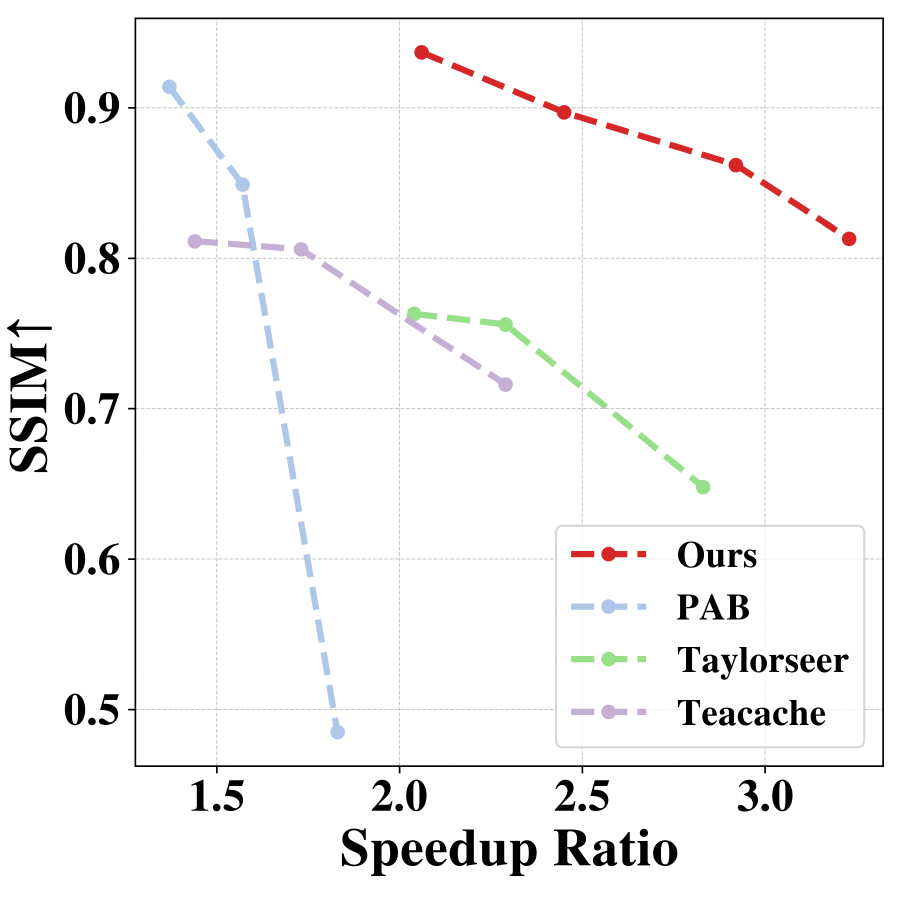}
    \hfill
    \includegraphics[width=0.32\textwidth]{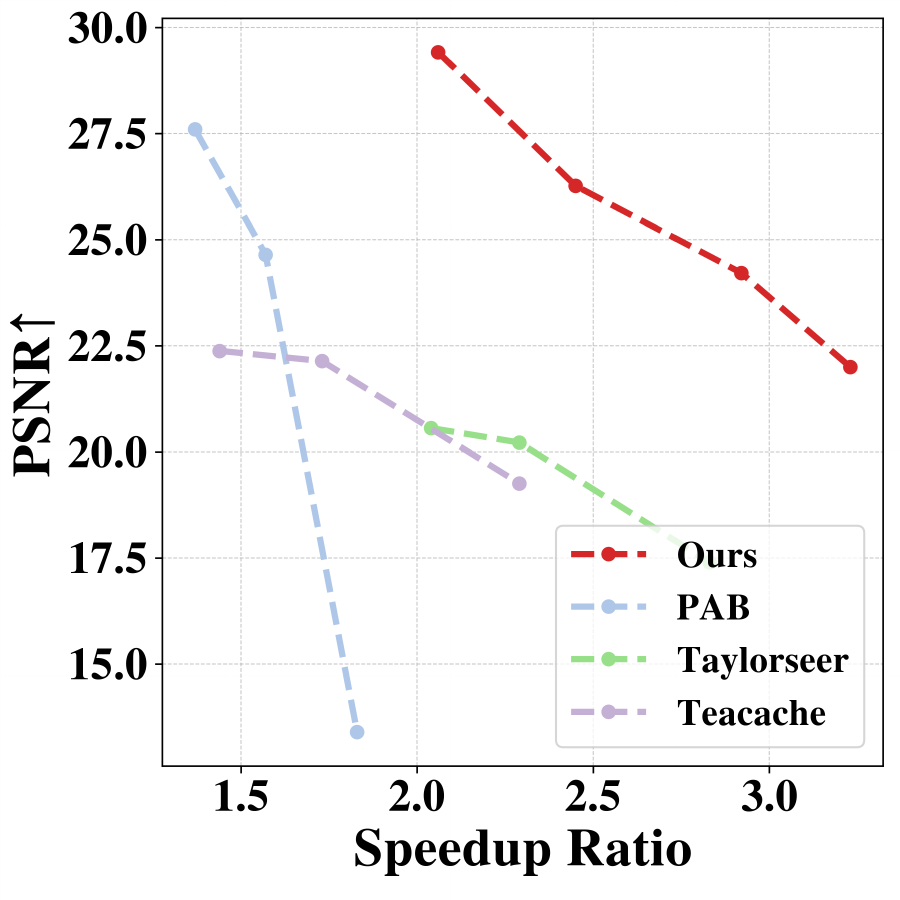}
    \hfill
    \includegraphics[width=0.32\textwidth]{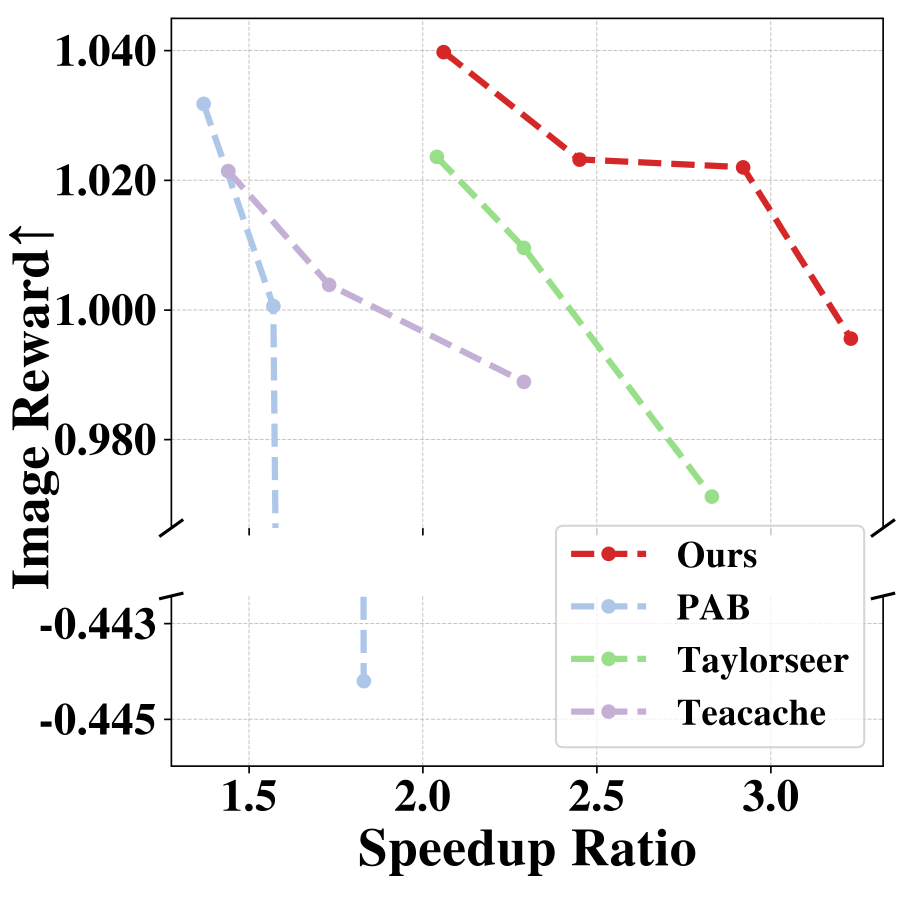}
    \vspace{-0.8em}
    \captionof{figure}{Comparison of our method and baselines under different speedup ratios.}
    \label{fig:speedratio}
      
\end{center}
}]
\footnotetext[1]{Corresponding author.}

\begin{abstract}
Diffusion Transformers (DiTs) have demonstrated remarkable performance in visual generation tasks. However, their low inference speed limits their deployment in low-resource applications. Recent training-free approaches exploit the redundancy of features across timesteps by caching and reusing past representations to accelerate inference. Building on this idea, TaylorSeer instead uses cached features to predict future ones via Taylor expansion. However, its module-level prediction across all transformer blocks (e.g., attention or feedforward modules) requires storing fine-grained intermediate features, leading to notable memory and computation overhead.  Moreover, it adopts a fixed caching schedule without considering the varying accuracy of predictions across timesteps, which can lead to degraded outputs when prediction fails. To address these limitations, we propose a novel approach to better leverage Taylor-based acceleration. First, we shift the Taylor prediction target from the module level to the last block level, significantly reducing the number of cached features. Furthermore, observing strong sequential dependencies among Transformer blocks, we propose to use the error between the Taylor-estimated and actual outputs of the first block as an indicator of prediction reliability. If the error is small, we trust the Taylor prediction for the last block; otherwise, we fall back to full computation, thereby enabling a dynamic caching mechanism.
Empirical results show that our method achieves a better balance between speed and quality, achieving a 3.17x acceleration on FLUX, 2.36x on DiT, and 4.14x on Wan Video with negligible quality drop. \redv{The Project Page is \href{https://cg-taylor-acce.github.io/CG-Taylor/}{here.}}
\end{abstract}    
\section{Introduction}
\begin{figure*}[t]
  \includegraphics[width=\textwidth]{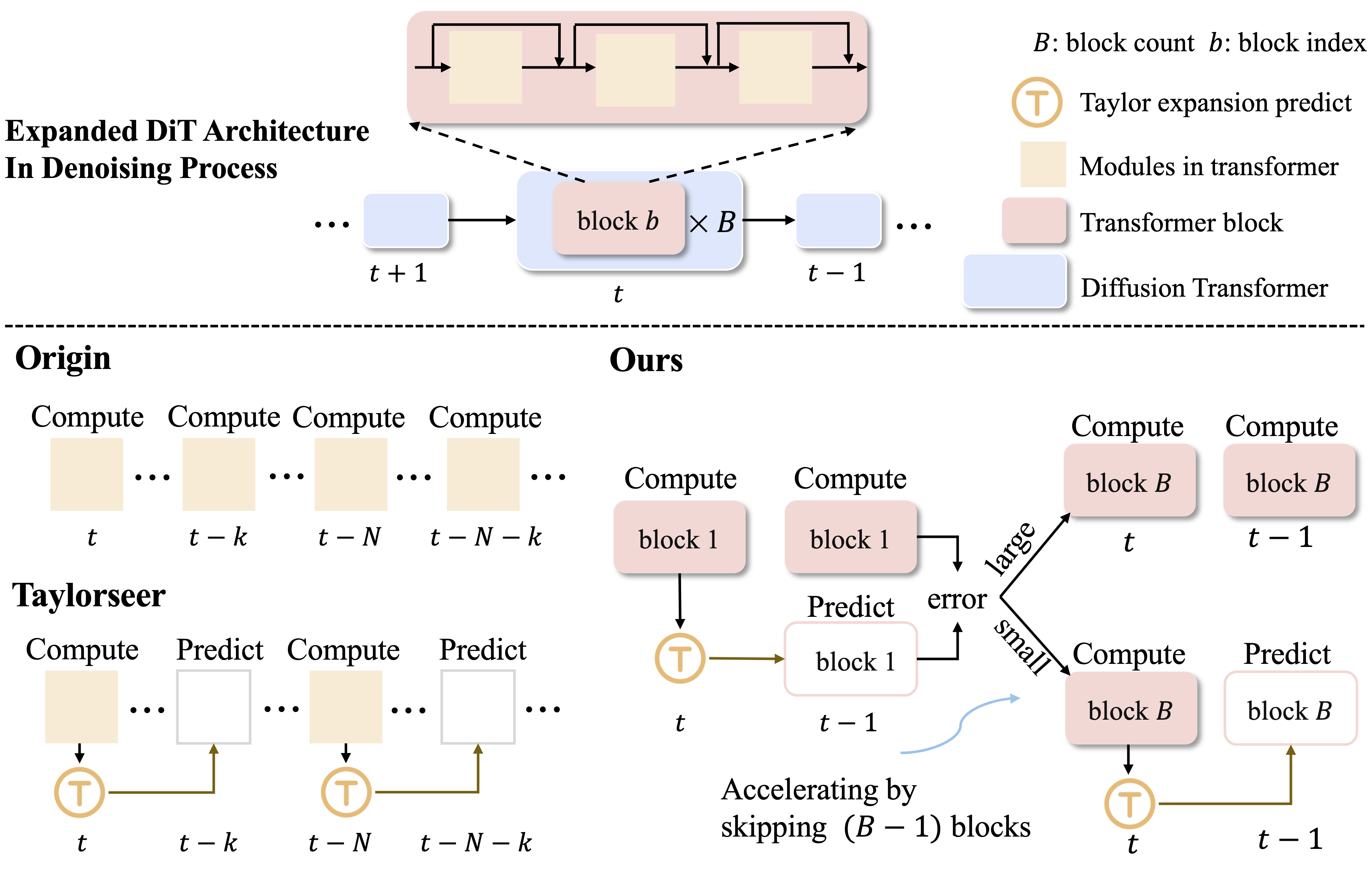}
  \caption{The expanded architecture of the DiT and the difference between different methods. (a) DiTs follow a hierarchical architecture which are composed of multiple transformer blocks, where each block typically consists of three modules. (b) In the original method, every module is computed at each timestep, resulting in a high computational cost. TaylorSeer reduced computation by calculating a module at one timestep and using Taylor expansions to predict its features for the remaining timesteps within a given interval $N$. Our method uses a confidence-gated method: if the first block’s prediction error at timestep t-1 is small,  we directly predict the last block’s feature at this timestep, skipping computation of the $B-1$ blocks to achieve acceleration. Otherwise, we compute all blocks normally.} 
  \vspace{-0.6cm}
  \label{fig:dit}

\end{figure*}
In recent years, diffusion models ~\cite{ho2020denoising,sohl2015deep,song2019generative,ke2024text,liu2025diffosr,qu2024mtlsc} have become foundational to visual generation, owing to their impressive ability to produce high-quality and semantically rich images. Initially built upon U-Net~\cite{ronneberger2015u} architectures, these models have demonstrated strong performance across domains, including text-to-image generation, inpainting, and video synthesis ~\cite{blattmann2023stable,ramesh2022hierarchical,li2025denoising,wang2025cymodiff,yuan2025diff,hu2025dtiu}. More recently, the introduction of Diffusion Transformers (DiT)~\cite{peebles2023scalable} has further advanced scalability and performance, enabling diffusion models to tackle complex generative scenarios and large-scale data~\cite{zheng2024open,yang2024cogvideox,zhuang2025enhanced,wang2025diff,zeng2025enhancing}. 
However, despite these advantages, the inherently iterative nature of the denoising process results in low inference speed. This limitation poses a significant challenge for deploying such models in low-resource or latency-sensitive applications.
To address the inefficiency of diffusion inference, a variety of approaches have been proposed, reducing the number of sampling steps, such as designing advanced probabilistic-flow-inspired solvers based on Ordinary Differential Equations (ODEs)~\cite{lu2022dpm,lu2025dpm,zheng2023dpm}. Other approaches focus on compressing model size by leveraging techniques like network pruning~\cite{zhu2024dip}, knowledge distillation~\cite{li2023snapfusion}, and token reduction~\cite{kim2024token,cheng2025cat}. However, most of these methods require additional training or offline optimization, introducing extra computational costs and limiting the overall deployment flexibility.

A recent line of work~\cite{chen2024delta,ma2024deepcache,selvaraju2024fora,zou2024accelerating,zou2024acceleratingduca} explores the feature redundancy inherent in diffusion models. Based on the observation that features across adjacent steps often exhibit high similarity, these approaches aim to accelerate inference by reusing features from previous timesteps. 
\red{Diffusion Transformers (DiTs) follow a hierarchical architecture as illustrated in Figure~\ref{fig:dit}. They are composed of multiple transformer blocks, where each block typically consists of three modules: cross-attention, self-attention, and feed-forward networks. During inference, every timestep must pass through all these modules, leading to substantial computational overhead. Intuitively, if the model could reuse the intermediate representations computed in earlier timesteps, rather than recalculating them from scratch at each timestep, the inference process would become significantly more efficient.}
By caching and reusing past representations instead of recomputing them, these methods reduce redundant computation and achieve faster sampling without requiring any additional training effort.
Nevertheless, as the gap between timesteps increases, the assumption of feature similarity gradually breaks down. Naively reusing cached features across distant steps can result in substantial errors, ultimately degrading image quality. To mitigate this, TaylorSeer~\cite{liu2025reusing} proposed to predict features at future timesteps using Taylor series expansion based on previously cached features, rather than directly reusing them. This enables the model to retain high generation quality, even under aggressive acceleration settings.

\red{While TaylorSeer improved the reliability of cached features by predicting future representations through Taylor expansion, it introduces considerable memory overhead and additional latency. Specifically, it applied Taylor expansion at every module within each transformer block (e.g., attention or feedforward module) as shown in Figure~\ref{fig:dit}, predicting the residual branch in a step-by-step manner.} However, modern diffusion transformer architectures consist of numerous modules, each requiring its own cached features. This design significantly increases memory consumption and computational burden, partially offsetting the intended acceleration gains.
To address this issue, we propose a \textbf{Last Block Forecast} strategy that reduces memory usage with minimal impact on acceleration performance. TaylorSeer revealed that intermediate features across blocks are highly predictable. Given the strictly sequential nature of DiT, where each block passes information to the next, we propose to forecast only the last block’s output, which corresponds to the model output at the current timestep. By doing so, we bypass the computations of the intermediate blocks to accelerate inference, while still avoiding the need to cache every module’s residual branch. This strategy modifies TaylorSeer’s original acceleration mechanism of forecasting instead of reuse, and it requires storing only a single set of features, thereby eliminating much of the computational overhead.

\red{Meanwhile, TaylorSeer adopted a fixed caching schedule by dividing the inference process into equally spaced intervals. As shown in Figure~\ref{fig:dit}, given an interval ${t,t-1,...,t-N+1}$, TaylorSeer will compute the feature at t and predict the rest of the timesteps using Taylor expansion.} This design implicitly assumes that Taylor prediction remains consistently reliable across timesteps, which often does not hold. In practice, the model’s output exhibits dynamic changes, leading to fluctuations in approximation accuracy over time. As a result, a rigid reuse pattern may either miss opportunities for further acceleration or incur significant errors when prediction fails. This limitation raises a natural and important question: \textit{When is the Taylor predictor reliable enough to replace actual computation?}

Intuitively, we only want to rely on the Taylor predictor when it is sufficiently accurate; otherwise, we fall back to the standard full computation. However, evaluating the full computation output just to compare with the predicted result defeats the purpose of acceleration.
To address this, we propose a \textbf{Prediction Confidence Gating (PCG)} mechanism. Our core hypothesis is that Transformer-based architectures like DiT exhibit strong sequential dependencies. In other words, the quality of early block predictions can reflect the predictability of later ones. Good predictions on early blocks suggest high reliability when forecasting the last block output.
Based on this insight, we use the prediction error of the first block as a proxy to assess the confidence of the Taylor prediction at the current timestep. Concretely, we compute the first block’s true output at each timestep and apply Taylor expansion to predict it in parallel. If the prediction error with respect to the ground truth is sufficiently small, we consider the model state to be sufficiently stable and apply the Taylor approximation for the remaining computation. Otherwise, we fall back to full computation to avoid quality degradation. This approach maintains decision quality while incurring almost no additional computational cost, enabling a training-free, dynamic acceleration strategy.

\vspace{-0.04cm}
Our method achieves up to a 3.17x acceleration on the FLUX model, 4.14x speedup on Wan Video, and 2.36x acceleration on DiT, with negligible degradation in image quality. Moreover, compared to TaylorSeer, our approach improves SSIM by approximately 25.5\% while being over one second faster, demonstrating a significantly better trade-off between efficiency and fidelity, as shown in Figure~\ref {fig:speedratio}.
\section{Related Work}
\subsection{Diffusion Transformer}
With the rapid development of deep learning~\cite{zhao2024mixir,liu2023reliable,qian2025visible}, diffusion models~\cite{ho2020denoising,sohl2015deep,liu2024iterative,wang2024silent,huang2025implicit} have achieved remarkable success in content generation, producing high-quality and diverse outputs. Early approaches~\cite{ramesh2022hierarchical,rombach2022high} were predominantly based on U-Net architectures, which delivered strong performance in both image and video generation. However, their scalability is fundamentally limited, posing challenges for building large-scale, high-capacity models.
To address this limitation, the Diffusion Transformer (DiT)~\cite{peebles2023scalable} architecture was proposed, leveraging the scalability and modeling flexibility of Transformers. This shift has led to a wave of progress across various generative tasks. Nonetheless, the broader adoption of DiT remains constrained by the inherently iterative nature of the diffusion process, which significantly hampers inference efficiency.

\subsection{Diffusion Model Acceleration}
With the growing deployment of diffusion models in real-world applications, improving their inference efficiency has become an increasingly important research direction. A variety of methods have been proposed to accelerate the sampling process, which can generally be categorized into \red{four} main paradigms.

First, several works aim to shorten the sampling trajectory~\cite{song2023consistency}. For example, methods like DDIM~\cite{song2020denoising} reformulate the generative process to avoid long iterative updates, while others draw inspiration from numerical solvers to make the generation process more efficient.
Second, model compression and quantization techniques have been explored to reduce computational overhead, including knowledge distillation~\cite{li2023snapfusion}, token reduction~\cite{kim2024token,cheng2025cat}, and network pruning~\cite{zhu2024dip,zhu2025dtsformer}. 
\red{Third, a set of training-oriented architecture modification methods aims to improve efficiency by altering the model design during training. For example, the Transformer architecture can be adapted to Mamba architecture~\cite{huang2025m4v}, attention modules can be replaced with Mamba attention~\cite{gao2024matten}, new modules can be introduced within the Transformer structure~\cite{po2025long}, or the Transformer can be designed without residual prediction~\cite{jiao2025flexvar} to reduce inference cost.} However, these methods often require additional training or offline optimization, which introduces extra computational cost.
Fourth, a growing body of work focuses on training-free acceleration via feature reuse~\cite{chen2024delta,ma2024deepcache,selvaraju2024fora,zou2024accelerating,zou2024acceleratingduca}, which has gained increasing attention due to its simplicity and effectiveness. These methods are motivated by the observation that intermediate features in diffusion models often contain redundancy, making it feasible to reuse them rather than recomputing at every step. For instance, PAB~\cite{zhao2024real} and T-Gate~\cite{liu2024faster} selectively reuse cached features based on different types of attention. TeaCache~\cite{liu2025timestep} proposes an adaptive caching strategy based on timestep-aware differences in output features. TaylorSeer~\cite{liu2025reusing} reduces the accumulation of cache reuse errors at distant timesteps by using Taylor expansion to predict feature values.
Building upon TaylorSeer, our method further improves prediction utility by explicitly evaluating the reliability of the Taylor approximation at each step. This enables a more principled decision on the predictability of a timestep, leading to a better trade-off between generation quality and inference efficiency.

\section{Method}

\subsection{Preliminary}
\textbf{Diffusion Models.}
Diffusion models generate data by learning to reverse a predefined noising process. The forward process gradually corrupts a clean sample $\mathbf{x}_0$ into a sequence of noisy latents $\mathbf{x}_1, \dots, \mathbf{x}_T$ by adding Gaussian noise at each timestep. The forward process is defined as:
\begin{equation}
q(\mathbf{x}_t \mid \mathbf{x}_0) = \mathcal{N}(\mathbf{x}_t; \sqrt{{\alpha}_t} \, \mathbf{x}_0, (1 - {\alpha}_t)\mathbf{I}),
\end{equation}

where ${\alpha}_t \in [0,1]$ represents the noise level. This formulation allows sampling $\mathbf{x}_t$ directly from $\mathbf{x}_0$ without computing intermediate steps. To generate new samples, the reverse process aims to recover $\mathbf{x}_0$ from pure noise $\mathbf{x}_T$ by iteratively denoising. This is achieved by training a neural network $\epsilon_\theta(\mathbf{x}_t, t)$ to predict the noise added at each timestep. Given this prediction, an estimate of the original image can be recovered by:

\begin{equation}
\hat{\mathbf{x}}_0 = \frac{1}{\sqrt{{\alpha}_t}} \left( \mathbf{x}_t - \sqrt{1 - {\alpha}_t} \, \epsilon_\theta(\mathbf{x}_t, t) \right).
\end{equation}

By reversing the noising process across all timesteps, the model gradually transforms random noise into a clean sample. This framework forms the basis for many recent advances in generative modeling across images, video, and beyond. However, this iterative denoising process incurs significant computational cost during inference, making real-time generation challenging.
\begin{figure*}[t]
  \includegraphics[width=\textwidth]{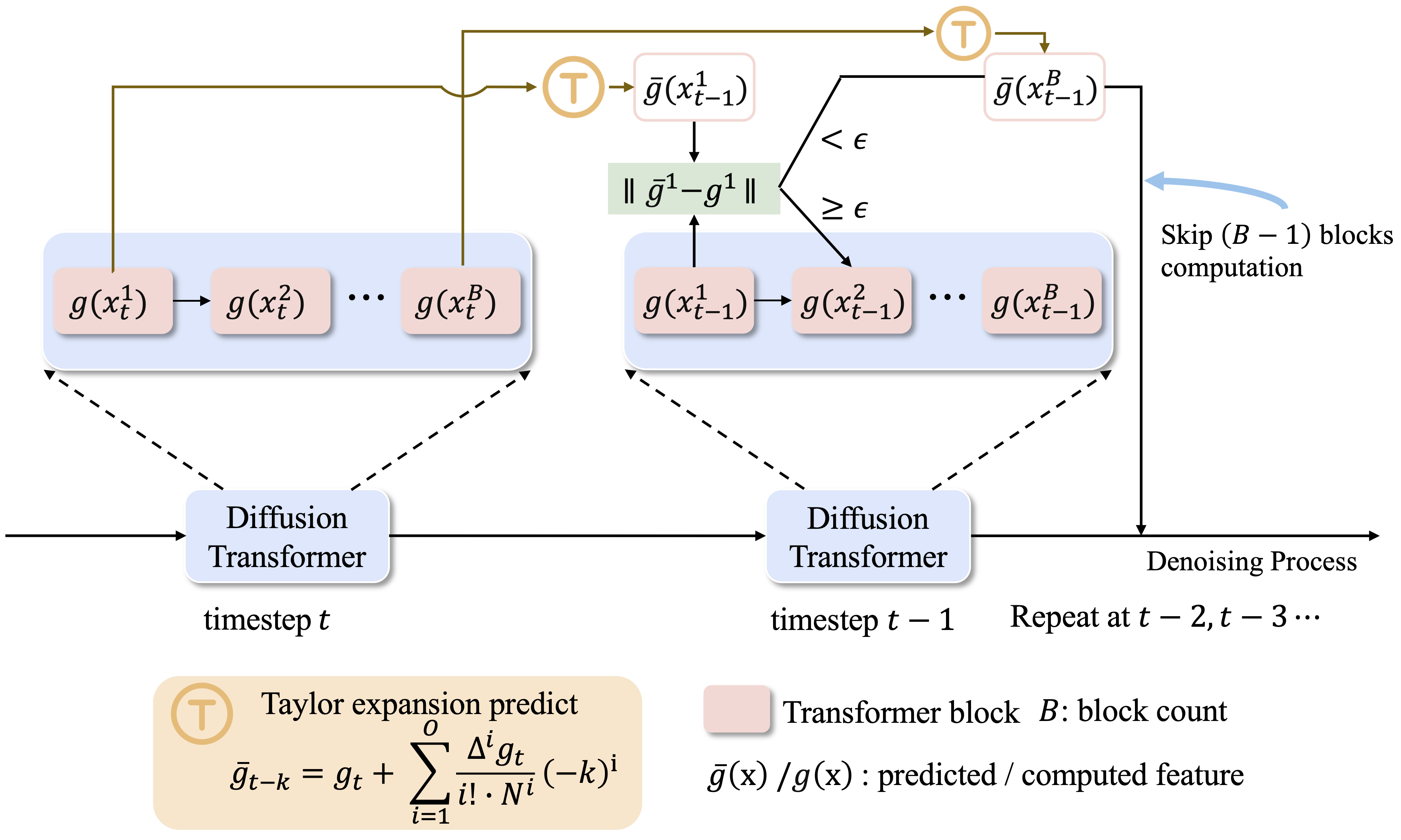}
  \caption{\red{The framework of our method. At timestep $t-1$, we first compute the actual output of the first block and simultaneously predict it using Taylor expansion with the cached feature at timestep $t$. The prediction error is then evaluated. If the error is below a threshold $\epsilon$, the Taylor prediction is considered reliable. In this case, we use it to approximate the last block feature, thereby skipping the computation of the remaining $(B-1)$ blocks to accelerate inference. Otherwise, we fall back to full computation of all blocks. This process is repeated for each subsequent timestep.}}
  \label{fig:method}
\end{figure*}
\textbf{Diffusion Transformer.}
The Diffusion Transformer (DiT) is a recently proposed architecture that replaces the traditional U-Net backbone with a Transformer-based design, offering improved scalability and expressiveness. A typical DiT model consists of a sequence of Transformer blocks $\{g^b\}_{b=1}^B$, where $B$ is the number
of blocks. Each Transformer block $g^b$ can be expressed as a composition of three modules:
\begin{equation}
g^b = \mathcal{F}_{SA}^{b} \circ \mathcal{F}_{CA}^{b}\circ \mathcal{F}_{FF}^{b},
\end{equation}
where $\mathcal{F}_{SA}^{b}$, $\mathcal{F}_{CA}^{b}$, and $\mathcal{F}_{FF}^{b}$ denote the self-attention, cross-attention, and feed-forward submodules in the $b$-th block, respectively. 
As the model progresses through the reverse process, these features are continuously updated, reflecting the model's step-wise refinement of the generated sample.

\textbf{TaylorSeer.} Previous feature caching methods accelerate inference by reusing the same features across an interval of timesteps. Specifically, given an interval $\{t, t{-}1, \dots, t{-}N{+}1\}$, they compute $\mathcal{F}(\mathbf{x}_t)$ once and approximate the rest as:
$\mathcal{F}(\mathbf{x}_{t-k}) \approx \mathcal{F}(\mathbf{x}_t), \text{for } k = 1, \dots, N{-}1.$
This direct reuse strategy, while efficient, leads to increasing approximation error as the time gap widens. To address this, TaylorSeer~\cite{liu2025reusing} proposed to predict features rather than reuse them verbatim.
Motivated by the observation that features and their derivatives evolve smoothly over timesteps, TaylorSeer applies Taylor series expansions using cached multi-order differences to capture the temporal dynamics of features. This method enables more accurate estimation of future features in a non-parametric manner.
At timestep $t$, they defined a cache storing the feature and its $O$-th order finite differences:
\begin{equation}
\mathcal{C}(x_t^b):=\{\mathcal{F}(x_t^b),\Delta\mathcal{F}(x_t^b),\dots \Delta^{O}\mathcal{F}(x_t^b)\},
\end{equation}
where $\mathcal{F}(x_t^b)$ denotes features at fully activated timesteps and $\Delta^i\mathcal{F}(x_t^b)$ represents the $i$-th order finite difference. To avoid explicit computation, they approximated higher-order derivatives using finite differences. Then the features for the $t-k$ steps are predicted as:
\begin{equation}
    \overline{\mathcal{F}}{(x_{t-k}^b)} = \mathcal{F}(x_t^b) + \sum_{i=1}^{O}\frac{\Delta^i\mathcal{F}(x_t^b)}{i!\cdot N^i}(-k)^i .
\end{equation}
\subsection{Last Block Forecast}
In TaylorSeer, each prediction unit is defined at the module level. That is, for every module within each transformer block, TaylorSeer stores intermediate features and their finite differences during inference to perform Taylor expansion. However, modern DiT architectures typically contain a large number of modules—often exceeding 100—resulting in substantial computational and memory overhead. Specifically, storing and predicting features for all modules incurs a cost proportional to almost $3 \times B \times O$. 

\red{TaylorSeer revealed that intermediate features
across blocks are highly predictable. Given the strictly sequential nature of DiT, where each block passes information to the next, we think that it is reasonable and feasible to accelerate the process by predicting the output of the last block. In this way, we can bypass the computations of the intermediate blocks while still obtaining the DiT output at the current timestep.}  Based on this, we propose a \textbf{Last Block Forecast} strategy: instead of applying the Taylor predictor to every module's intermediate features, we apply it directly to the last block at each timestep.
Formally, given the last block feature $g(x_t^B)$ at timestep $t$, we cache the following features and finite differences:
\begin{equation}
\mathcal{C}(x_t^B):=\{{g}(x_t^B),\Delta{g}(x_t^B),\dots \Delta^{O}{g}(x_t^B)\}.
\end{equation}
Then, we estimate future outputs for the $t-k$ step using a Taylor expansion:

\begin{equation}
    \overline{g}{(x_{t-k}^B)} = {g}(x_t^B) + \sum_{i=1}^{O}\frac{\Delta^i{g}(x_t^B)}{i!\cdot N^i}(-k)^i.
\label{eq:taylorseer_lastBlock}
\end{equation}

The Last Block Forecast strategy reduces the caching requirement from $3 \times B \times O$ features to just $O$, significantly lowering both memory usage and prediction time. Moreover, it enables efficient acceleration while preserving the benefits of higher-order approximation, making it particularly suitable for large-scale DiT models.
\subsection{Prediction Confidence Gating}
While the last block forecast reduces the overhead of feature caching and prediction, the design where TaylorSeer adopted a fixed caching schedule during inference can still lead to quality degradation, especially when the predicted feature significantly deviates from the true feature. Therefore, it is critical to determine \textit{when} to trust the Taylor predictor and \textit{when} to fall back to actual computation.

We discover that Transformer-based architectures like DiT, the blocks exhibit strong sequential
dependencies—meaning the quality of early block predictions often reflects the predictability of later blocks. Motivated by this insight, we introduce a lightweight proxy estimation method, Prediction Confidence Gating (PCG), which uses the prediction error of the first block to estimate the reliability of Taylor-based prediction at the current timestep (Figure~\ref{fig:method}). Specifically, at each timestep $t-k$, we apply Taylor expansion to predict the first Transformer block output, using the same expansion form defined in Eq.~\ref{eq:taylorseer_lastBlock}, but applied at the first block:

\begin{equation}
    \overline{g}{(x_{t-k}^1)} = {g}(x_t^i) + \sum_{i=1}^{O}\frac{\Delta^i{g}(x_t^1)}{i!\cdot N^i}(-k)^i.
\label{eq:output_taylorseer}
\end{equation}

Meanwhile, we also compute the full output of the first block, denoted as ${g}(x_{t-k}^1)$, at each timestep. We compare this output with the corresponding Taylor-predicted result using a predefined error metric: 
\begin{equation}
\frac{
\left\| \overline{g}{(x_{t-k}^1)}  -g{(x_{t-k}^1)}  \right\|}{\left\|{g}{(x_{t-k}^1)}\right\|} < \epsilon.
\label{eq:firstblock-threshold}
\end{equation}
If the approximation error is below a threshold $\epsilon$, we consider the Taylor expansion reliable at this timestep. In this case, we allow the Taylor-predicted features to estimate the output of the last transformer block. Otherwise, we fall back to full computation:
This strategy allows us to dynamically adapt the use of Taylor-based prediction, ensuring that inference speedup does not come at the cost of significant quality degradation. Since the first block is shallow and fast to compute, the overhead of this additional check is minimal. 

\section{Experiments}
\subsection{Experiment Settings}
\begin{table*}[]
\centering
\caption{Quantitative comparison of our method and baselines on FLUX for text-to-image generation.}
\label{tab:flux}
\resizebox{\textwidth}{!}{%
\begin{tabular}{@{}l|cccccc@{}}
\toprule
\textbf{Method} & \multicolumn{3}{c}{\textbf{Acceleration}} & \multicolumn{3}{c}{\textbf{Visual   Quality}} \\ \cmidrule(l){2-7} 
\textbf{FLUX.1~\cite{flux}} & \textbf{FLOPs(T)↓} & \textbf{Latency(s)↓} & \multicolumn{1}{c|}{\textbf{Speed↑}} & \multicolumn{1}{c|}{\textbf{Image Reward↑}} & \textbf{SSIM ↑} & \textbf{PSNR↑} \\ \midrule
dev 50steps & 1910.10 & 23.23 & \multicolumn{1}{c|}{1.00x} & \multicolumn{1}{c|}{1.0398} & - & - \\ \midrule
PAB~\cite{zhao2024real} & 1169.60 & 14.80 & \multicolumn{1}{c|}{1.57x} & \multicolumn{1}{c|}{1.0006} & 0.849 & 24.65 \\
Teacache ($\delta= 0.25$)~\cite{liu2025timestep} & 1069.84 & 13.55 & \multicolumn{1}{c|}{1.71x} & \multicolumn{1}{c|}{1.0039} & 0.806 & 22.13 \\
Taylorseer ($N$ = 2, $O$ = 1)~\cite{liu2025reusing} & 993.53 & 13.56 & \multicolumn{1}{c|}{1.71x} & \multicolumn{1}{c|}{1.0005} & 0.850 & 24.02 \\
Ours ($\epsilon$ = 0.03) & \textbf{826.82} & \textbf{11.26} & \multicolumn{1}{c|}{\textbf{2.06x}} & \multicolumn{1}{c|}{\textbf{1.0398}} & \textbf{0.937} & \textbf{29.42} \\ \midrule
Teacache ($\delta$ = 0.40)~\cite{liu2025timestep} & 802.48 & 10.14 & \multicolumn{1}{c|}{2.29x} & \multicolumn{1}{c|}{0.9889} & 0.716 & 19.26 \\
Taylorseer ($N$ = 5, $O$ = 2)~\cite{liu2025reusing} & \textbf{458.86} & 8.21 & \multicolumn{1}{c|}{2.83x} & \multicolumn{1}{c|}{0.9793} & 0.648 & 17.40 \\
Ours ($\epsilon$ = 0.13) & 503.72 & \textbf{7.19} & \multicolumn{1}{c|}{\textbf{3.13x}} & \multicolumn{1}{c|}{\textbf{1.0216}} & \textbf{0.813} & \textbf{22.00} \\ \bottomrule
\end{tabular}%
}

\end{table*}

\begin{table*}[]
\centering
\caption{Quantitative comparison of our method and baselines on the FLUX model for text-to-image generation on the GenEval benchmark.}
\label{tab:geneval}
\resizebox{\textwidth}{!}{%
\begin{tabular}{@{}l|c|ccccccc@{}}
\toprule
Method & \multicolumn{1}{l|}{\textbf{Acceleration}} & \multicolumn{7}{c}{\textbf{GenEval}} \\ \cmidrule(l){2-9} 
\textbf{FLUX.1~\cite{flux}} & \textbf{Speed↑} & \multicolumn{1}{c|}{\textbf{overall↑}} & \textbf{Single Obj.↑} & \textbf{Two Obj.↑} & \textbf{Counting↑} & \textbf{Colors↑} & \textbf{Position↑} & \textbf{Attr. Binding↑} \\ \midrule
dev 50steps & 1.00x & \multicolumn{1}{c|}{0.67} & 1.00 & 0.83 & 0.73 & 0.78 & 0.21 & 0.48 \\ \midrule
PAB~\cite{zhao2024real} & 1.57x & \multicolumn{1}{c|}{0.66} & 0.99 & 0.82 & 0.72 & 0.76 & \textbf{0.21} & 0.47 \\
Teacache ($\delta$ = 0.25)~\cite{liu2025timestep} & 1.71x & \multicolumn{1}{c|}{0.66} & 0.99 & 0.81 & \textbf{0.73} & 0.78 & 0.18 & 0.46 \\
Taylorseer($N = 2, O = 1$)~\cite{liu2025reusing} & 1.71x & \multicolumn{1}{c|}{\textbf{0.67}} & 0.99 & 0.83 & \textbf{0.73} & \textbf{0.79} & 0.20 & \textbf{0.49} \\
Ours($\epsilon$ = 0.03) & \textbf{2.06x} & \multicolumn{1}{c|}{\textbf{0.67}} & \textbf{1.00} & \textbf{0.84} & \textbf{0.73} & 0.77 & 0.20 & 0.48 \\ \midrule
Teacache ($\delta$ = 0.40)~\cite{liu2025timestep} & 2.29x & \multicolumn{1}{c|}{\textbf{0.67}} & \textbf{1.00} & 0.83 & 0.72 & \textbf{0.79} & 0.19 & 0.47 \\
Taylorseer($N = 5, O = 2$)~\cite{liu2025reusing} & 2.83x & \multicolumn{1}{c|}{0.65} & 0.99 & \textbf{0.84} & 0.71 & 0.45 & \textbf{0.21} & 0.45 \\
Ours($\epsilon$ = 0.13) & \textbf{3.13x} & \multicolumn{1}{c|}{\textbf{0.67}} & 0.99 & 0.81 & \textbf{0.77} & 0.77 & \textbf{0.21} & \textbf{0.48} \\ \bottomrule
\end{tabular}%
}
\end{table*}

 \textbf{Model Configurations.} 
 To validate the generality of our method across diverse generative settings, we evaluate it on \red{four} representative diffusion models spanning different modalities and resolutions. FLUX.1-dev~\cite{flux}, a text-to-image model, produces high-resolution outputs of 1024×1024 using the Rectified Flow sampler~\cite{liu2022flow}. DiT-XL/2~\cite{peebles2023scalable}, a class-conditional image generator, follows a 50-step DDIM~\cite{song2020denoising} schedule to generate 256×256 images. Meanwhile, Wan Video~\cite{wan2025wan}, designed for text-to-video generation, synthesizes 81-frame clips at 480×832 resolution, adhering to its original sampling configuration. \red{Hunyuan Video~\cite{kong2024hunyuanvideo} also designed for text-to-video generation, generates 61-frame clips at 320×512 resolution using the default diffusers implementation, with the only modification being that we set the inference steps to 50.} This diverse selection of models ensures a comprehensive evaluation of both inference efficiency and generation quality.

FLUX, Wan Video, and Hunyuan Video models are implemented in the \redv{PaddlePaddle}~\cite{ma2019paddlepaddle} framework, while DiT is implemented using PyTorch~\cite{paszke2019pytorch}.
To ensure fairness and compatibility, our acceleration method is integrated into each model’s original inference pipeline without altering its underlying architecture or training setup.
All experiments are conducted on NVIDIA A800 80GB GPUs, and latency is measured under consistent hardware and software conditions. 

 \textbf{Evaluation and Metrics.}
We evaluate the effectiveness of our method from two perspectives: acceleration and visual quality. Our experimental setup closely follows the configurations of TaylorSeer~\cite{liu2025reusing} and TeaCache~\cite{liu2025timestep} to ensure fair comparison.
 To assess inference efficiency, we uniformly report inference latency and Floating-Point Operations (FLOPs) as the primary metrics. 
For generation quality, we incorporate task-specific evaluation metrics tailored to each generation setting. For text-to-image generation, we adopt the DrawBench benchmark~\cite{saharia2022photorealistic} and evaluate the results using ImageReward~\cite{xu2023imagereward}, which estimates alignment with human preferences. \red{In addition, we employ GenEval~\cite{ghosh2023geneval}, an object-focused framework to evaluate compositional image properties across multiple dimensions.} Furthermore, we use PSNR and SSIM to quantify the similarity between outputs produced by our accelerated method and those generated by the original models for text-to-image generation on the COCO-1K dataset. For class-conditional image generation, we employ the FID-50k~\cite{heusel2017gans}, sFID-50k, and Inception Score, which measures the distributional similarity between generated samples and real training data. For text-to-video generation, we use VBench~\cite{huang2024vbench}, a video-specific benchmark containing 946 prompts, designed to reflect human-aligned quality judgments. In addition to Vbench, we further report PSNR, SSIM, and LPIPS~\cite{zhang2018unreasonable} to assess the perceptual and structural similarity between videos generated by different methods. 

\begin{figure}[t]
  \centering
   \includegraphics[width=\linewidth]{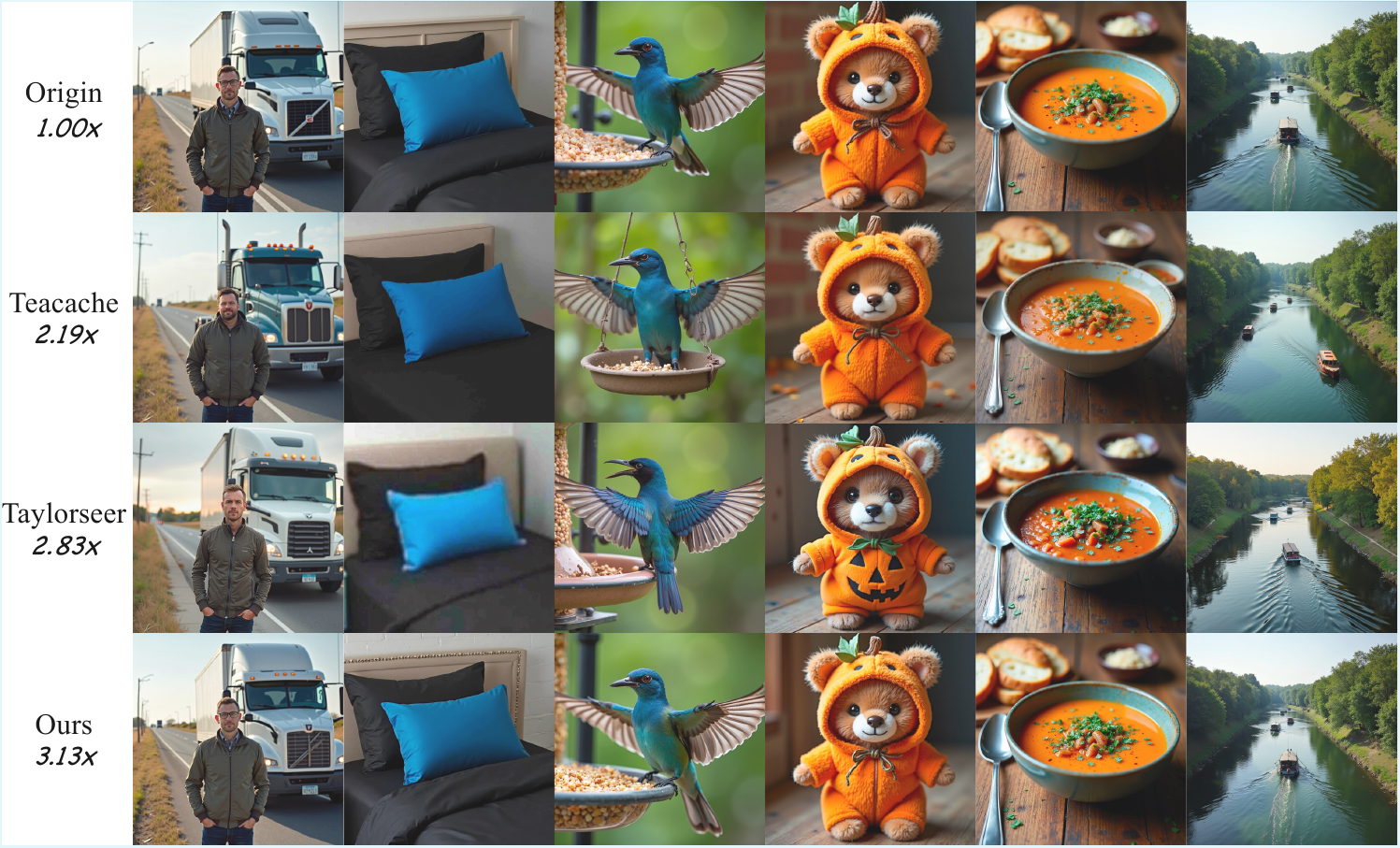} 
   \caption{
 Visualization of different acceleration methods on FLUX. Our method achieves higher visual quality and greater similarity to the original image while operating at a faster speed.
   }
   \label{fig:flux_vis}
\end{figure}

\begin{table*}[]
\centering
\caption{Quantitative comparison of our method and baselines on the DiT model for class-to-image generation.}
\label{tab:dit}
\resizebox{\textwidth}{!}{
\begin{tabular}{@{}l|c|ccc|ccc@{}}
\toprule
\textbf{Method} & \textbf{Effcient} & \multicolumn{3}{c|}{\textbf{Acceleration}} & \multicolumn{3}{c}{\textbf{Visual Quality}} \\ \cmidrule(l){3-8} 
\textbf{DiT~\cite{peebles2023scalable}} & \textbf{Attention~\cite{dao2022flashattention}} & \textbf{FLOPs(T)↓} & \textbf{Latency(s)↓} & \textbf{Speed↑} & \textbf{FID↓} & \multicolumn{1}{c|}{\textbf{sFID ↓}} & \textbf{Inception Score↑} \\ \midrule
ori 50steps & \ding{51} & 23.74 & 0.428 & 1.00x & 2.32 & \multicolumn{1}{c|}{4.32} & 241.25 \\ \midrule
ori 20 steps & \ding{51} & 9.49 & 0.191 & 2.24x & 3.18 & \multicolumn{1}{c|}{5.15} & 221.43 \\
FORA ($N=3$)~\cite{selvaraju2024fora} & \ding{51} & 8.58 & 0.197 & 2.17x & 3.55 & \multicolumn{1}{c|}{6.36} & 229.02 \\
ToCa ($N=3$)~\cite{zou2024accelerating} & \ding{55} & 10.23 & 0.216 & 1.98x & 2.87 & \multicolumn{1}{c|}{4.76} & 235.21 \\
DuCa ($N=3$)~\cite{zou2024acceleratingduca} & \ding{51} & 9.58 & 0.208 & 2.06x & 2.88 & \multicolumn{1}{c|}{4.66} & 233.37 \\
Taylorseer ($N=3, O=3$)~\cite{liu2025reusing} & \ding{51} & 8.56 & 0.248 & 1.73x & 2.34 & \multicolumn{1}{c|}{4.69} & \textbf{238.42} \\
Ours ($\epsilon=0.08$) & \ding{51} & 12.86 & \textbf{0.181} & \textbf{2.36x} & \textbf{2.34} & \multicolumn{1}{c|}{\textbf{4.28}} & 235.76 \\ \bottomrule
\end{tabular}
}
\end{table*}

\begin{table*}[]
\centering
\caption{Quantitative comparison of our method and baselines on Wan Video for text-to-video generation.}
\label{tab:wan}
\resizebox{\textwidth}{!}{%
\begin{tabular}{@{}l|cc|cccc@{}}
\toprule
\textbf{Method} & \multicolumn{2}{c|}{\textbf{Acceleration}} & \multicolumn{4}{c}{\textbf{Visual Quality}} \\ \midrule
\textbf{Wan Video~\cite{wan2025wan}} & \textbf{Latency(s)↓} & \textbf{Speed↑} & \textbf{VBench(\%)↑} & \textbf{SSIM ↑} & \textbf{PSNR↑} & \textbf{LPIPS ↓} \\ \midrule
ori 50steps & 165.69 & 1.00x & 82.72 & - & - & - \\ \midrule
PAB~\cite{zhao2024real} & 78.02 & 2.12x & 73.60 & 0.5205 & 17.1515 & 0.4783 \\
Teacache ($\delta=0.26$)~\cite{liu2025timestep} & 43.25 & 3.83x & 76.67 & 0.4147 & 12.4317 & 0.6332 \\
Taylorseer ($N=5, O=1$)~\cite{liu2025reusing} & 45.27 & 3.66x & 78.71 & 0.4167 & 12.6739 & 0.5758 \\
Ours ($\epsilon=0.36$) & \textbf{40.02} & \textbf{4.14x} & \textbf{81.86} & \textbf{0.6319} & \textbf{17.2108} & \textbf{0.2882} \\ \bottomrule
\end{tabular}%
}
\end{table*}


\begin{table*}[]
\centering
\caption{Quantitative comparison of our method and baselines on Hunyuan Video for text-to-video generation.}
\label{tab:hunyuan}
\resizebox{\textwidth}{!}{%
\begin{tabular}{@{}l|cc|cccc@{}}
\toprule
\textbf{Method} & \multicolumn{2}{c|}{\textbf{Acceleration}} & \multicolumn{4}{c}{\textbf{Visual Quality}} \\ \midrule
\textbf{Hunyuan Video~\cite{kong2024hunyuanvideo}} & \textbf{Latency(s)↓} & \textbf{Speed↑} & \textbf{VBench(\%)↑} & \textbf{SSIM ↑} & \textbf{PSNR↑} & \textbf{LPIPS ↓} \\ \midrule
ori 50steps & 101.48 & 1.00x & 80.18 & - & - & - \\ \midrule
PAB~\cite{zhao2024real} & 47.35 & 2.14x & 79.16 & 0.6193 & 16.8222 & 0.3707 \\
Teacache ($\delta=0.15$)~\cite{liu2025timestep} & 40.55 & 2.50x & 79.71 & 0.6602 & 18.1942 & 0.3049 \\
Taylorseer ($N=5, O=1$)~\cite{liu2025reusing} & 30.20 & 3.36x & 78.88 & 0.5765 & 15.4932 & 0.4309 \\
Ours ($\epsilon=0.12$) & \textbf{27.45} & \textbf{3.70x} & \textbf{79.75} & \textbf{0.7287} & \textbf{20.5806} & \textbf{0.2159} \\ \bottomrule
\end{tabular}%
}
\end{table*}

\subsection{Text-to-image model}
\textbf{Quantitative Study.}
We evaluate our method under two different settings and compare it with prior works, including PAB~\cite{zhao2024real}, Teacache~\cite{liu2025timestep}, and TaylorSeer~\cite{liu2025reusing}. As shown in Table~\ref{tab:flux}, our approach achieves the best trade-off between inference efficiency and generation quality in both settings. Under the low acceleration setting, our method achieves the fastest latency of 11.26s, representing a 2.06× speedup, while also attaining the highest Image Reward (1.0398) and SSIM (0.937), with PSNR (29.42)—substantially outperforming other baselines. Importantly, under this acceleration setting, the Image Reward score remains identical to that of the original model, demonstrating that our method preserves generation quality without compromise.
In the more aggressive setting, our method further reduces latency to 7.19s, with a 3.13× speedup, maintaining competitive visual quality (Image Reward 0.9956, SSIM 0.813, PSNR 22.00). Compared to TaylorSeer~\cite{liu2025reusing}, our method achieves a latency reduction of 1.02 seconds, while simultaneously improving SSIM by 25.5\% and PSNR by 26.4\%.

\red{Furthermore, we evaluate the GenEval benchmark on the FLUX model. The results are shown in Table~\ref{tab:geneval}. It demonstrates that different methods achieve relatively close performance on GenEval.  However, our approach consistently demonstrates strong competitiveness under both low and high acceleration ratios. This further validates the effectiveness of our method.}

\textbf{Qualitative Study.}
Figure \ref{fig:flux_vis} shows that our method preserves high image quality even under accelerated inference. The generated images retain the essential subjects and background elements from the original outputs. In contrast, other methods often yield blurry results or common-sense inconsistencies, such as lamps appearing on rooftops, an incomplete truck in Teacache~\cite{liu2025timestep}, or a fuzzy bed in Taylorseer~\cite{liu2025reusing}. These results highlight our method’s ability to achieve a better trade-off between inference speed and visual fidelity.

\subsection{Class-to-Image model}
\textbf{Quantitative Study.}
To further validate the generality of our method, we conduct experiments on DiT using various baseline acceleration strategies. As shown in Table \ref{tab:dit}, our method offers a superior balance between inference speed and visual quality.
Specifically, our method attains the fastest inference latency (0.181s) and the highest speed-up ratio (2.36×) while preserving visual quality comparable to the original 50-step generation. In terms of quality metrics, we achieve an FID of 2.34, sFID of 4.28, outperforming most efficient baselines such as FORA~\cite{selvaraju2024fora}, ToCa~\cite{zou2024accelerating}, and DuCa~\cite{zou2024acceleratingduca}.
Notably, although TaylorSeer achieves a slightly lower FLOPs count, it suffers from higher latency due to its module-wise prediction overhead. In contrast, our method applies a lightweight last block forecast and a dynamic gating mechanism, resulting in lower runtime overhead without sacrificing fidelity.
In addition, our method preserves efficient attention mechanisms~\cite{dao2022flashattention} throughout, ensuring compatibility with common lightweight DiT backbones.
\begin{table*}[t]
\centering
\caption{Performance comparison under different prediction confidence thresholds $\epsilon$.}
\label{tab:thresold}
\resizebox{\textwidth}{!}{%
\begin{tabular}{@{}c|cc|ccc@{}}
\toprule
\textbf{Confidence Thresholds $\epsilon$} & \textbf{Latency(s)↓} & \textbf{Speed↑} & \textbf{Image Reward↑} & \textbf{SSIM↑} & \textbf{PSNR↑} \\ \midrule
0.03 & 11.26 & 2.06x & 1.0398 & 0.937 & 29.42 \\
0.08 & 7.95 & 2.92x & 1.0220 & 0.862 & 24.22 \\
0.13 & 7.19 & 3.23x & 1.0216 & 0.813 & 22.00 \\
0.18 & 6.13 & 3.79x & 0.9994 & 0.773 & 20.81 \\ \bottomrule
\end{tabular}%
}
\end{table*}

\subsection{Text-to-Video model}
\textbf{Quantitative Study.}
\red{Quantitative results on the Wan Video model are shown in Table~\ref{tab:wan}. Our method achieves the highest inference speed, with a latency of only 40.02 seconds, representing a 4.14× acceleration over the original 50-step baseline (165.69s). Notably, this speedup outperforms all other training-free acceleration methods, including Teacache (3.83×)~\cite{liu2025timestep} and Taylorseer (3.66×)~\cite{liu2025reusing}, while also offering superior visual quality. Specifically, our approach maintains a VBench score of 81.86\%, just 0.9\% below the original unaccelerated baseline (82.72\%), yet significantly higher than PAB (73.60\%)~\cite{zhao2024real}, Teacache (76.67\%)~\cite{liu2025timestep}, and TaylorSeer (78.71\%)~\cite{liu2025reusing}. In terms of perceptual quality, our method achieves the best SSIM (0.6319), PSNR (17.2108), and LPIPS (0.2882), which outperform TaylorSeer by 34.06\%, 26.36\%, and 49.94\%, respectively, demonstrating that our adaptive caching mechanism helps reduce approximation errors and better preserve visual details.
These results suggest that our method not only enables the highest acceleration but also minimizes quality degradation, effectively balancing speed and fidelity.}

\red{Quantitative results on the Hunyuan Video model are shown in Table~\ref{tab:hunyuan}. Consistent with the results on Wan Video, our method achieves the fastest inference speed on Hunyuan Video while maintaining high visual quality. Specifically, it reaches a 3.70× speedup with an inference latency of 27.45s, outperforming baselines such as TaylorSeer, TeaCache, and PAB. The VBench score drops minimally compared to the original model, and perceptual metrics (SSIM, PSNR, LPIPS) indicate that our approach also provides superior visual fidelity, with significant improvements over TaylorSeer. Overall, experiments on these two text-to-video models further validate the effectiveness and generalizability of our method. }

\subsection{Ablation Study}

\begin{figure}[t]
  \centering
   \includegraphics[width=0.5\linewidth]{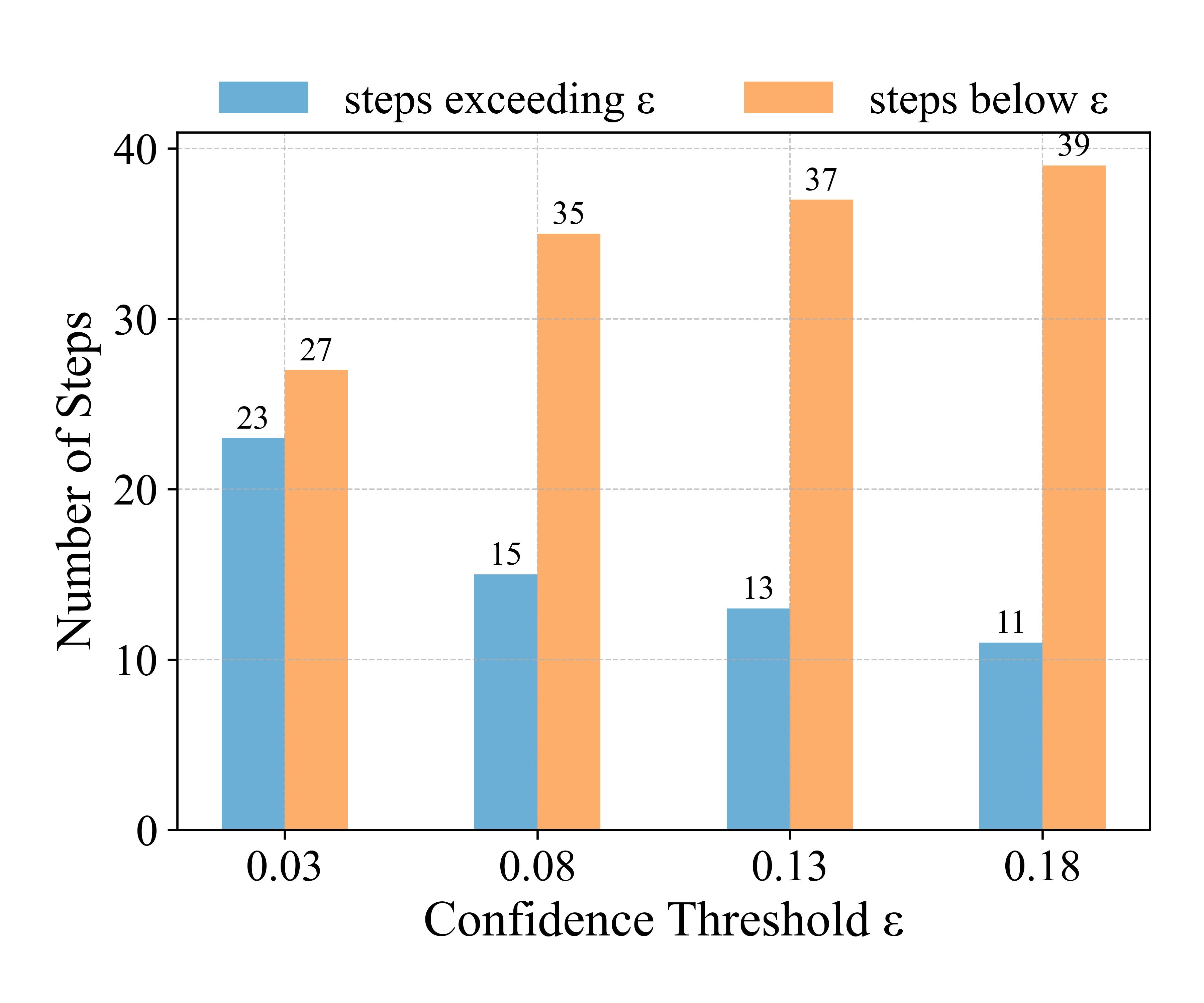} 
   \caption{
   \red{Number of denoising steps below and above the $\epsilon$ threshold for different $\epsilon$ values.}}
   \label{fig:exceed}
\end{figure}

\textbf{Performance of Different Prediction Confidence Thresholds $\epsilon$:}
\red{Table~\ref{tab:thresold} plots the speed–quality trade‑off of our method as the confidence threshold $\epsilon$ varies ($\epsilon \in \red{\{0.03,0.08,0.13,0.18\}}$). As expected, larger thresholds admit more aggressive reuse and thus yield higher speed‑up ratios, while a smaller $\epsilon$ results in slower inference but better preservation of image quality. This controllable trade-off allows us to select manually a balanced $\epsilon$ that achieves both substantial acceleration and high output fidelity. Furthermore, to better illustrate how $\epsilon$ influences acceleration, we report Figure~\ref{fig:exceed} showing the number of steps exceeding or falling below the $\epsilon$ threshold at different values. As shown in the figure, with a larger $\epsilon$, more steps fall below the threshold, which means that during inference, a greater portion of steps rely on Taylor-predicted features of the last block. Consequently, the acceleration effect becomes more pronounced.} Although image‑quality metrics (Image Reward, SSIM, PSNR) gradually decline with increasing $\epsilon$, our curves remain above those of all baselines at comparable speeds, indicating a slower quality‑decay rate as is shown in Figure~\ref{fig:speedratio}. The advantage is most pronounced relative to PAB: once PAB approaches a 2× speed‑up, its quality drops sharply, whereas our method still delivers substantially higher fidelity. These results confirm that the Prediction Confidence Gating mechanism provides a flexible knob for trading speed for quality while preserving a consistently better balance than competing approaches.

\begin{table*}[t]
\centering
\caption{Performance comparison between constant and timestep-dependent thresholds $\epsilon$.}
\label{tab:ada_threshold}
\resizebox{\textwidth}{!}{%
\begin{tabular}{@{}c|cc|ccc@{}}
\toprule
\textbf{Confidence Thresholds $\epsilon$} & \textbf{Latency(s)↓} & \textbf{Speed↑} & \textbf{Image Reward↑} & \textbf{SSIM↑} & \textbf{PSNR↑} \\ \midrule
0.08 & 7.95 & 2.92x & 1.0220 & 0.862 & 24.22 \\
adaptive $\epsilon$ & 8.11 & 2.86x & 1.0339 & 0.904 & 26.91 \\ \bottomrule
\end{tabular}%
}
\end{table*}

\begin{table*}[]
\centering
\caption{Ablation study of each component in our method on FLUX for text-to-image generation.}
\label{tab:compo}
\resizebox{\textwidth}{!}{%
\begin{tabular}{@{}cc|c|ccc|c@{}}
\toprule
\textbf{Forecast level} & \textbf{PCG} & \textbf{Latency(s)↓} & \textbf{Image Reward↑} & \textbf{PSNR↑} & \textbf{SSIM ↑} & \textbf{Memory Usage (GB)↓} \\ \midrule
module & \ding{55} & 13.56 & 1.0005 & 24.02 & 0.850 & 39.69 \\
last block & \ding{55} & 12.41 & 1.0109 & 24.00 & 0.849 & \textbf{35.73} \\
last block & \ding{51} & \textbf{11.26} & \textbf{1.0398} & \textbf{29.42} & \textbf{0.937} & 35.78 \\ \bottomrule
\end{tabular}%
}
\end{table*}
\begin{table*}[t]
\centering
\caption{Performance comparison when different blocks are used as prediction indicators.}
\label{tab:prediction}
\resizebox{\textwidth}{!}{%
\begin{tabular}{@{}c|cc|ccc@{}}
\toprule
\textbf{Prediction Indicator} & \textbf{Latency(s)↓} & \textbf{Speed↑} & \textbf{Image Reward↑} & \textbf{SSIM↑} & \textbf{PSNR↑} \\ \midrule
First Block & 7.19 & 2.92x & 1.0217 & 0.813 & 22.00 \\
Third Block & 7.49 & 2.86x & 1.0170 & 0.801 & 21.77 \\ \bottomrule
\end{tabular}%
}
\end{table*}
\red{In addition, motivated by~\cite{wu2024diffusion}, we also experimented with a timestep-dependent $\epsilon$. $\epsilon$ is set smaller at early timesteps and larger at later ones, since early denoising determines global structure while later steps mainly refine details. As reported in Table ~\ref{tab:ada_threshold}, the timestep-dependent $\epsilon$ schedule achieves better trade-offs. At nearly identical inference speeds (only 0.16s difference), it improves image reward, SSIM, and PSNR, with PSNR increased by 11.2\%. This demonstrates that our framework is flexible and can naturally benefit from adaptive thresholding strategies. While our main method adopts a constant $\epsilon$ for simplicity and clarity, we believe that designing principled adaptive schemes is a promising future direction.}

\textbf{Effectiveness of each Component.} 
To evaluate the effectiveness of each component in our method, we conduct an ablation study focusing on two key modules: Last Block Forecast and Prediction Confidence Gating (PCG), as shown in Table~\ref{tab:compo}.
To ensure fairness, we keep the TaylorSeer-related configurations consistent across all variants, applying the same Taylor order and caching intervals when applicable.

When the forecast level is set to module, the method corresponds to the original TaylorSeer. Switching to the Last Block Forecast reduces latency from 13.56s to 12.41s, while maintaining nearly the same visual quality—SSIM drops only marginally from 0.850 to 0.849, and PSNR from 24.02 to 24.00. Notably, ImageReward slightly improves from 1.0005 to 1.0109, suggesting that our Last Block Forecast effectively eliminates redundant intermediate predictions without degrading generation quality. With both the Last Block Forecast and PCG enabled, the visual quality improves substantially. Despite achieving even faster inference (latency reduced by 1s compared to Last Block Forecast-only), SSIM rises to 0.937 and PSNR reaches 29.42—a relative improvement of 22.58\%. ImageReward also increases from 1.0109 to 1.0398. These results demonstrate that PCG enables a better trade-off between speed and image quality. We also evaluate the peak GPU memory usage during inference. Our method reduces GPU memory consumption by approximately 10\% (from 39.69 GB to 35.78 GB), which is particularly beneficial for large-scale DiT models.

\begin{figure}[t]
  \centering
   \includegraphics[width=0.7\linewidth]{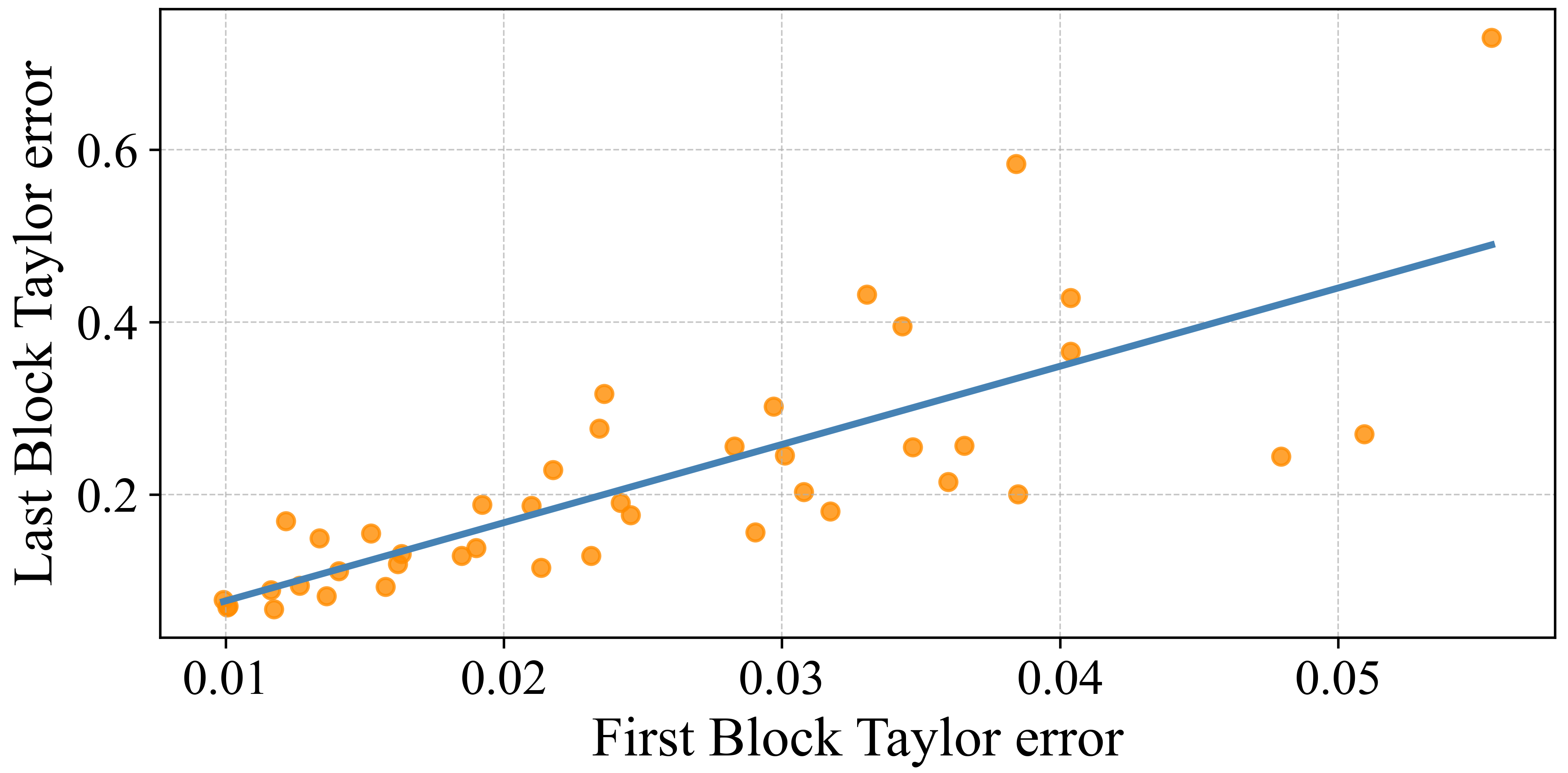} 
   \caption{Correlation between the first block and the last block of Taylor errors.}
   \label{fig:error_relate}
\end{figure}

\textbf{Analysis of the Prediction Confidence Gating.}
To examine whether the first‑block error is a reliable proxy for the accuracy of the last-block Taylor prediction, we plot the per‑timestep errors defined in Eq.\ref{eq:firstblock-threshold} in Figure \ref{fig:error_relate}. The scatter plot reveals a clear positive correlation: when the Taylor error of the first block is small, the error of the last block output is likewise small, and vice versa. This strong coupling confirms the effectiveness of our Prediction Confidence Gating (PCG). By invoking Taylor prediction only when the first‑block error is low—and reverting to standard computation otherwise—PCG exploits Taylor’s speed advantage whenever it is trustworthy while sidestepping cases that would degrade quality.

\red{\textbf{Choice of Prediction Indicator.}}
\red{In our experiments, we use the first block as the indicator to assess the accuracy of Taylor prediction. The rationale is that the first block provides a lightweight signal, thereby minimizing the overhead for confidence evaluation. To validate this design choice, we further conduct an ablation study by selecting the third block as the prediction indicator. The results, reported in Table~\ref{tab:prediction}, show that the visual quality remains nearly unchanged compared to using the first block, with SSIM differing by only 0.13. However, the inference speed decreases by 4.17\%. This confirms that using the first block strikes the optimal balance—it serves as an effective predictor while incurring less runtime cost.}

\section{Conclusion}
We propose a training-free framework to accelerate diffusion model inference by leveraging the predictive power of Taylor expansion more effectively. Unlike prior methods that rely on fixed reuse patterns and module-level predictions, our approach introduces a last block forecast strategy to reduce memory and computation, and a prediction confidence gating mechanism that dynamically decides when to trust Taylor-based approximations. Experiments on multiple models demonstrate that our method achieves superior speed-quality trade-offs, offering faster generation with minimal quality degradation. We believe our findings shed light on the importance of dynamic control and last block approximation in efficient diffusion inference. This paves the way for acceleration strategies that are more practical and adaptive in real-world applications.

\section*{Acknowledgments}
This work was supported by the Key Research and Development Program of Yunnan Province under Grant No. 202403AA080002

{
    \small
    \bibliographystyle{ieeenat_fullname}
    \bibliography{main}
}


\end{document}